\documentclass[letterpaper, 10 pt, conference]{ieeeconf}  

\IEEEoverridecommandlockouts                              

\usepackage{graphicx}
\usepackage{url}
\usepackage{multicol}
\usepackage{graphicx}
\usepackage{hyperref}
\usepackage[export]{adjustbox}
\usepackage{subcaption}
\usepackage{wrapfig}

\usepackage{multicol}
\usepackage{booktabs}
\usepackage{algorithm}
\usepackage[noend]{algpseudocode}
\usepackage{float}
\usepackage{amsmath, amssymb}
\usepackage{amsfonts}
\usepackage[skip=0pt,font=small,labelfont=bf]{caption}
\usepackage{bm}
\usepackage{epstopdf}

\usepackage{color}
\definecolor{purple}{RGB}{210, 0, 210}
\definecolor{international_orange}{RGB}{240, 74, 0}

\usepackage{cite} %

\newcommand{\reals}{\mathbb{R}}
\renewcommand{\vec}{\mathbf}

\newcommand{\DeformerNet}{\emph{DeformerNet}}
\newcommand{\pcloud}{\mathcal{P}}
\newcommand{\curpcloud}{\pcloud_\mathrm{c}}
\newcommand{\goalpcloud}{\pcloud_\mathrm{g}}
\newcommand{\initpcloud}{\pcloud_\mathrm{i}}
\newcommand{\tpcloud}{\pcloud_t} %

\newcommand{\object}{\mathcal{O}}
\newcommand{\policy}{\pi}
\newcommand{\sspolicy}{\pi_\mathrm{s}}
\newcommand{\action}{\mathcal{A}}
\newcommand{\deltapose}{\Delta\vec{x}}

\newcommand{\matches}{\mathcal{M}}
\newcommand{\manippoint}{\vec{p}_\mathrm{m}}
\newcommand{\encoder}{g}
\newcommand{\feat}{\psi}
\newcommand{\featcur}{\psi_\mathrm{c}}
\newcommand{\featgoal}{\psi_\mathrm{g}}
\newcommand{\deltafeat}{\Delta \feat}
\newcommand{\deformfunc}{F}

\title{Learning Visual Shape Control of Novel 3D Deformable Objects \\from Partial-View Point Clouds}

\author{Bao Thach\(^1\), Brian Y. Cho\(^1\), Alan Kuntz\(^1\), Tucker Hermans\(^{1,2}\)
\thanks{$^{1}$Robotics Center and School of Computing, University of Utah, Salt Lake City, UT 84112, USA; $^{2}$NVIDIA Corporation, Seattle, WA, USA; {\tt\{bao.thach, brian.cho, alan.kuntz, tucker.hermans\}@utah.edu} B.T. was supported in part by NSF Award \#2024778.
}}
    
\begin{document}
\maketitle
\thispagestyle{empty}
\pagestyle{empty}

\begin{abstract}
If robots could reliably manipulate the shape of 3D deformable objects, they could find applications in fields ranging from home care to warehouse fulfillment to surgical assistance.
Analytic models of elastic, 3D deformable objects require numerous parameters to describe the potentially infinite degrees of freedom present in determining the object's shape.
Previous attempts at performing 3D shape control rely on hand-crafted features to represent the object shape and require training of object-specific control models.
We overcome these issues through the use of our novel \DeformerNet{} neural network architecture, which operates on a partial-view point cloud of the object being manipulated and a point cloud of the goal shape to learn a low-dimensional representation of the object shape.
This shape embedding enables the robot to learn to define a visual servo controller that provides Cartesian pose changes to the robot end-effector causing the object to deform towards its target shape. 
Crucially, we demonstrate both in simulation and on a physical robot that \DeformerNet{} reliably generalizes to object shapes and material stiffness not seen during training and outperforms comparison methods for both the generic shape control and the surgical task of retraction.
\end{abstract}

\section{Introduction}\label{sec:intro}
Manipulation of 3D deformable objects stands at the heart of many tasks we wish to assign to autonomous robots.
For example, home-assistance robots must be able to manipulate objects such as sponges, mops, bedding, and food to help people with day-to-day life. Robots operating in warehouses should safely handle deformable containers such as bags and boxes in order to package outgoing orders. Factory robots benefit from the ability to remove deformable objects from containers.
Most critically, surgical assistive robots are required to safely and precisely manipulate deformable tissue and organs.

However, 3D deformable object manipulation presents many challenges~\cite{Sanchez2018robotic}.
The shape of deformable objects require a potentially infinite number of degrees of freedom (DOF) to describe, compared to only 6 DOF for rigid objects.
As a result, deriving low-dimensional but accurate and expressive state representations for deformable objects is difficult. An additional challenge compared with simpler linear deformable objects such as ropes and cloth arises as elastic 3D deformable objects cannot be released without returning to their initial configuration.
Further, deformable objects frequently have complex dynamics~\cite{Wu2020Learning}, making the process of deriving a model laborious and potentially computationally intensive. These issues all present themselves in the specific problem we examine in this work: 3D deformable object shape control. The shape control problem requires a robot to manipulate the internal DOF of a 3D deformable object to reach a desired shape.

\begin{figure}[t]
  \centering
  \includegraphics[width=0.49\linewidth,clip,trim=0cm 10cm 10cm 25cm]{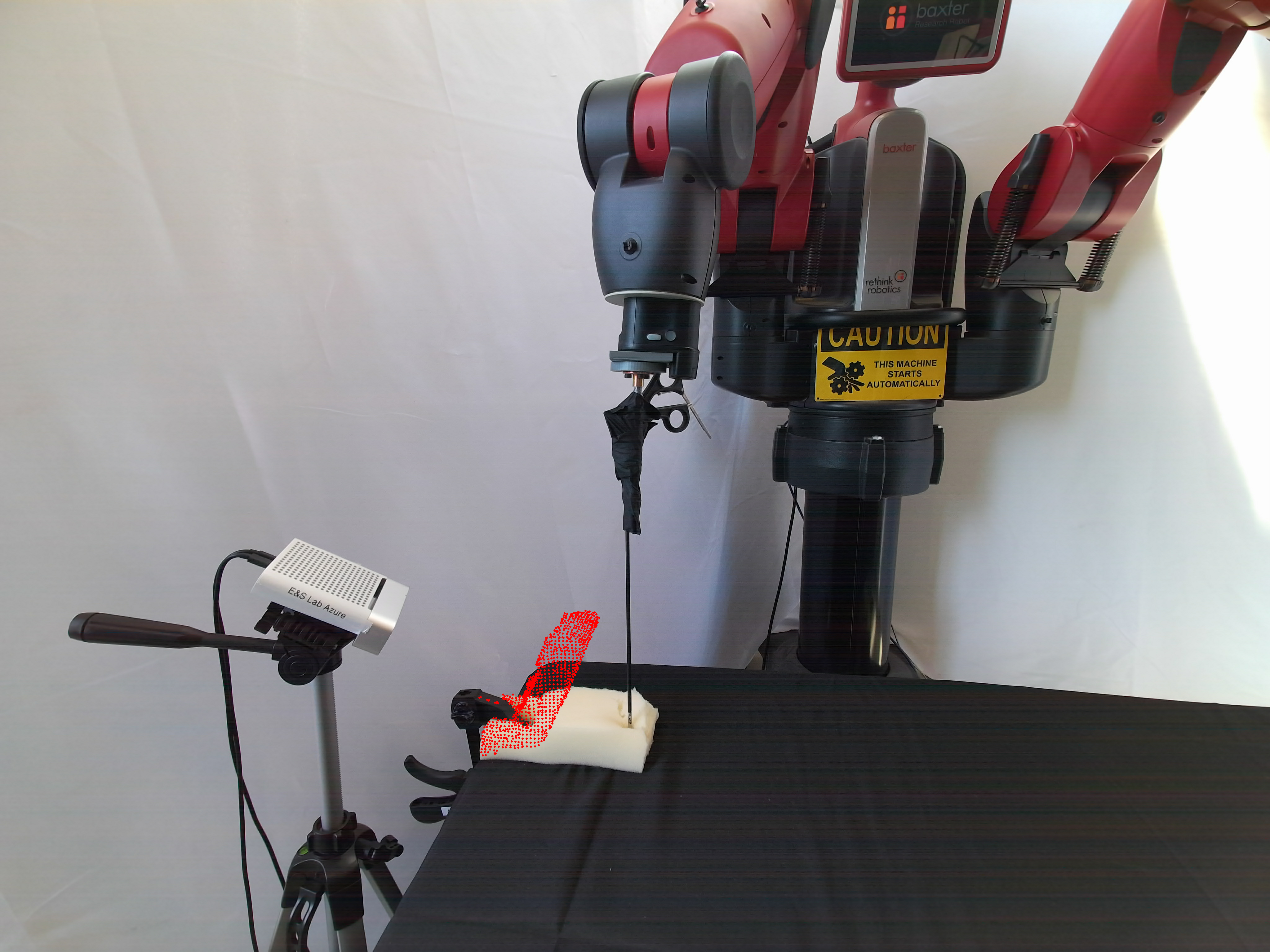}
  \includegraphics[width=0.49\linewidth,clip,trim=0cm 10cm 10cm 25cm]{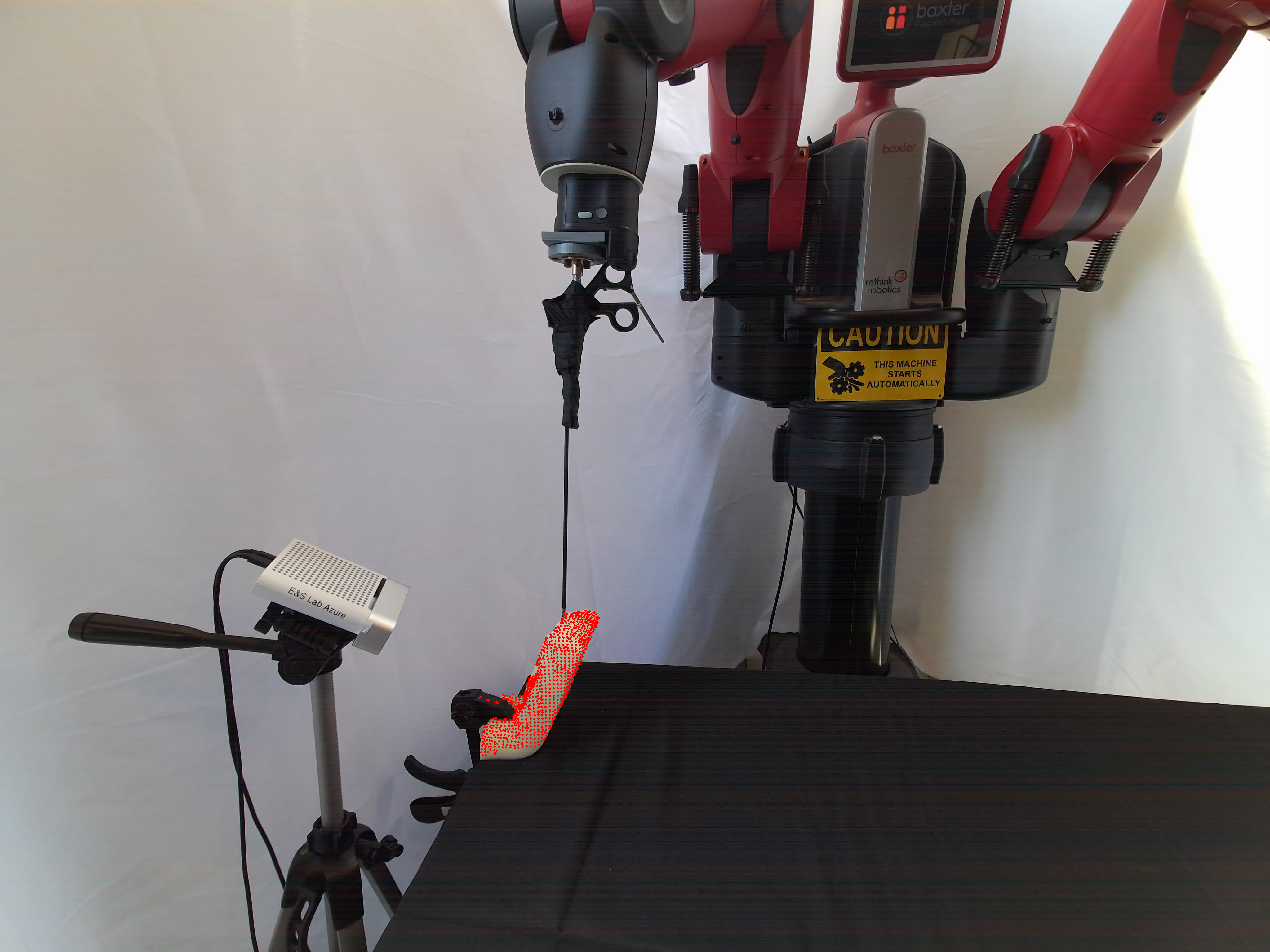}
    \caption{Example initial and final configurations of object shape control via shape servo with \DeformerNet{} on a physical robot using a laparoscopic tool. We visualize in red the goal point clouds given to the controller.}
    \label{fig:intro}
    \vspace{-20pt}
\end{figure}

While rigid-body manipulation has received a large amount of study~\cite{mason2018toward}, due to the challenges listed above, autonomous 3D deformable object manipulation currently still remains an under-researched area~\cite{Sanchez2018robotic,huang2021defgraspsim}---despite its potential relevance and need.
Existing work for 3D deformable shape control leverages hard-coded feature vectors to describe deformable object state~\cite{Hu20193-D}, which struggles to represent large sets of shapes.
While learning-based methods show great promise in both rigid~\cite{lu-ram2020-grasp-inference,mousavian2019graspnet} and deformable object manipulation~\cite{huang2021defgraspsim,Yan2020Learning}, these methods require a large amount of training data.
Due to the difficulty of accurately simulating deformable objects, existing methods for shape control rely on data gathered via real-world setups, limiting the efficacy of learning-based approaches.
Further, the ability to successfully manipulate deformable material is heavily dependent on where the robot grasps an object, however current works do not provide methods for selecting grasping points conditioned on the desired post-grasp manipulation.

In this work, we take steps toward addressing each of these gaps in the context of 3D deformable shape control.
Our method takes as input a partial-view point cloud representation of a 3D deformable object and a desired goal shape.
We build our method around a novel neural-network architecture, \DeformerNet{}, which is trained on a large amount of data gathered via a recently-developed high-fidelity deformable object simulator, Isaac Gym~\cite{Liang2018GPU,huang2021defgraspsim,macklin2019}.

Our method first reasons over the initial and target shape to select a manipulation point.  Following selection of this grasp point, \DeformerNet{} takes as input the current and target point clouds of the object, embeds the shape into a low-dimensional latent space representation, and computes a change in end-effector position that moves the object closer to the goal shape.
The robot executes this motion and proceeds in a closed-loop fashion generating commands from \DeformerNet{} until reaching the desired goal shape. Figure~\ref{fig:intro} shows the initial and final configurations from an example manipulation using \DeformerNet{} on a physical robot. Our results provide the first empirical demonstration of the importance of manipulation point selection for 3D shape control. 

We focus our evaluation on the surgical robotics domain.
We task a robot with manipulating three classes of object primitives into a variety of goal shapes using a laparoscopic tool. Unlike the preliminary results presented in our previous workshop paper~\cite{thach2021deformernet} we vary the dimensions and the stiffness properties of the objects.
We demonstrate effective manipulation on test objects both in simulation and on a physical robot. Importantly we show that our method can manipulate objects that fall both inside and outside the distributions of object shape and stiffness seen in training. We show that our DeformerNet outperforms both a sampling-based strategy and a model-free reinforcement learning approach on the shape control task.

We additionally present a strategy for applying our method to the common surgical task of retraction where we simplify the need of a target shape to only specifying a plane which the deformable tissue needs to be on one side of. We demonstrate successful retraction both in simulation and on the physical robot. We make available all code and data associated with this paper at \url{https://sites.google.com/view/deformernet/home}.

\section{Related Work}\label{sec:related_work}
Many approaches leverage machine learning with point cloud sensing to manipulate 3D rigid objects~\cite{murali2020dof,mousavian2019graspnet,deng2020self,lu-ram2020-grasp-inference,lu-iros2020-active-grasp,vandermerwe-icra2020-reconstruction-grasping}. Authors have proposed various neural network architectures to encode object shape to achieve varying tasks such as grasp planing~\cite{mousavian2019graspnet,lu-ram2020-grasp-inference,lu-iros2020-active-grasp,vandermerwe-icra2020-reconstruction-grasping}, collision checking~\cite{murali2020dof}, shape completion~\cite{vandermerwe-icra2020-reconstruction-grasping}, and object pose estimation~\cite{deng2020self}.
In this work, we build upon these concepts to apply a learning-based approach which reasons over point cloud sensing with learned feature vectors to manipulate 3D deformable objects.

Solutions to 3D deformable object shape control~\cite{Sanchez2018robotic} can be categorized into learning-based and learning-free approaches.
Among the learning-free methods, a series of papers~\cite{Navarro-Alarcon2013b, Navarro-Alarcon2014, Navarro-Alarcon2016} define a set of geometric features on the object as the state representation. The authors use this representation to perform visual servoing with adaptive linear controller.
These methods only work for known objects with distinct texture and cannot generalize to a diverse set of objects.
This formulation controls the displacements of individual points which does not fully reflect the 3D shape of the object.
For precise control, one must use a large number of feature points, making control highly susceptible to noise and occlusion.
Other learning-free works~\cite{Qi2019Contour,Navarro-Alarcon2017,zhu2021vision} represent the object shape using 2D image contours; this severely limits the space of controllable 3D deformations.

Among learning-based 3D shape control methods, Hu \emph{et al.}~\cite{Hu20193-D} represents the current state-of-the-art work in 3D shape control.
Specifically, they use extended FPFH \cite{Rusu2010VFH} to extract a feature vector from an input point cloud and learn to predict deformation actions via a neural network to control objects to desired shapes.
However, we show that this architecture over-simplifies the complex dynamics of 3D deformable objects and thus struggles to learn to control to a diverse set of target shapes~\cite{thach2021deformernet}.

There has also been work on shape control of deformable objects that exhibit lower dimensional behavior, e.g., 1D objects such as rope, and 2D objects, such as cloth~\cite{Wu2020Learning,matas2018sim,Yan2020Learning,ma2020contrast,McConachie2017bandit}. These methods typically either directly learn a policy using model-free RL that map
RGB images of the object to robot actions~\cite{Wu2020Learning, matas2018sim} or learn predictive models of the object under robot actions~\cite{Yan2020Learning,ma2020contrast,ma2021learning,McConachie2017bandit}.
These 1D and 2D works do not scale to the 3D deformable object shape control problem, either because they leverage lower dimensional object or sensing (e.g. RGB images) representation or the inherent physical differences between 1D, 2D, and 3D objects (e.g. 3D elastic tissue will return to its initial shape after released).

With respect to surgical robotics, several learning-based approaches have been applied to other surgical tasks including suturing~\cite{vandenberg2010_ICRA,Chiu2021Bimanual}, cutting~\cite{Thananjeyan2017_ICRA, Murali2015_ICRA}, tissue tracking~\cite{Lu2021Super}, and simulation~\cite{xu2021surol}.
In this work we apply our method to surgical retraction.
Attanasio et al.~\cite{Attanasio2020_RAL} propose the use of surgeon-derived heuristic motion primitives to move tissue flaps identified by a vision system.
In~\cite{Jansen2009_IROS}, a grasp location and planar retraction trajectory is computed with a linearized potential energy model leveraging online simulation.
In~\cite{meli2021autonomous}, a logic-based task planner is leveraged which guarantees interpretability, however this work focuses on manipulating a single thin tissue sheet and does not show shape or material property generalization or validation on a physical robot.
Nagy et al.~\cite{Nagy2018_SAMI} propose the use of stereo vison accompanied by multiple control methods, however the method assumes a thin tissue layer and a clear view of two tissue layers.
Pore et al.~\cite{pore2021safe} introduce a model-free reinforcement learning method which learns safe motions for a robot's end effector during retraction, however it does not explicitly reason over the deformation of the tissue.
We compare against a similar approach, using the same model-free reinforcement learning algorithm, but adapted to our task to explicitly reason over the tissue state.

\section{Problem Formulation}\label{sec:problem}
We address the problem of robotically manipulating a 3D deformable object from an initial shape to a goal shape.
In this context, \emph{3D} refers to \emph{triparametric} or \emph{volumetric} objects \cite{Sanchez2018robotic} which have no dimension significantly smaller than the other two, unlike \emph{uniparametric} (e.g., rope) and \emph{biparametric} objects (e.g., cloth).

We define the shape of the 3D volumetric object to be manipulated as $\object \subset \reals^3$, noting that it will change over time as the robot manipulates it and the object interacts with the environment.
As typical robots cannot directly sense $\object$, we consider a partial-view point cloud $\pcloud \subset \object$ as a subset of the points on the surface of $\object$, due to the prevalence of sensors that produce point clouds.
We define the point cloud representing the initial shape of the object as $\initpcloud$, the goal shape for the object as $\goalpcloud$, and the shape of the object at a given intermediate point in time $\curpcloud$.

\begin{figure}[ht]
    \centering
    \includegraphics[width=\linewidth]{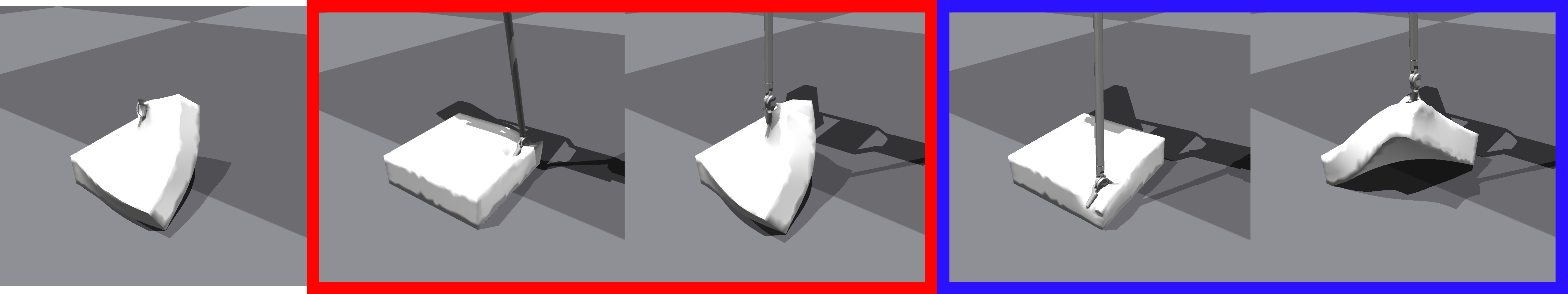} %
    \caption{Importance of manipulation point (MP) selection. Leftmost: goal shape; Red box: successful MP; Blue box: failed MP.}
    \label{fig:mani_point}
    \vspace{-12pt}
\end{figure}

We note that the successful manipulation of a deformable object depends on the point on the object the robot grasps, i.e., the manipulation point (see Fig.~\ref{fig:mani_point}).
As such, we present the first problem as the selection of a manipulation point, which we define as $\manippoint = [x, y, z] \in \object$.

Having grasped the object, the robot can change that object's shape by moving its end-effector and in turn moving the manipulation point of the object.
We define a manipulation action $\action$ as a change in the manipulation point, formally $\action \in \reals^3, \action = \deltapose = [\Delta x, \Delta y, \Delta z]$.
The resulting problem then becomes to define a policy  $\policy: \pcloud \times \pcloud \to \reals^3$, which maps the point cloud representing the object shape and the goal point cloud to an action vector describing the change in manipulation point that drives the object toward the goal shape, i.e., $\policy(\curpcloud, \goalpcloud) = \action$.
The repeated application of a successful policy $\policy$ results in a manipulation point trajectory, which when executed by the robot, results in transforming the object from its initial shape to a goal shape.

\section{Method}\label{sec:method}
In this section we explain the details of our proposed approach. We first explain our \textit{shape servo}~\cite{Navarro-Alarcon2017} approach to create a feedback policy for 3D deformable object shape control. Following this we give details of the \DeformerNet{} network architecture at the heart of our shape servo policy. Finally in this section we present our approach to selecting a manipulation point, conditioned on the goal configuration, used by the robot while performing shape control.

\subsection{Shape Servo Control with \DeformerNet{}}
The shape servo formulation~\cite{Navarro-Alarcon2017,Hu20193-D} uses visual feedback, here in the form of partial-view point clouds of the object being manipulated, as input to a policy that computes a robot action that attempts to instantaneously bring the current shape, \(\curpcloud\) closer to the target shape, \(\goalpcloud\).

\begin{figure}[ht]
    \centering
    \vspace{-10pt}
    \includegraphics[width=1\linewidth]{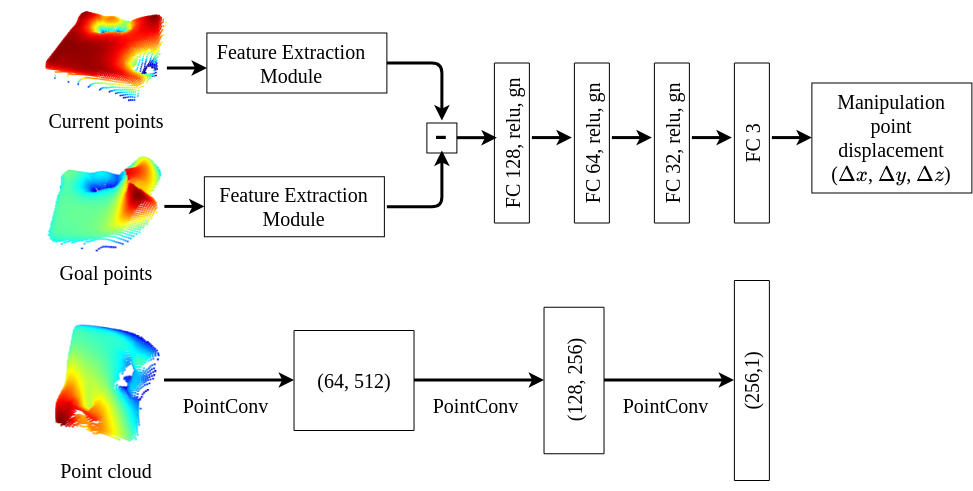}%
    \caption{(Top) Architecture of \DeformerNet{}; (Bottom) architecture of the feature extraction module.}
    \label{fig:DeformerNet}\vspace{-10pt}
\end{figure}

Following the notation from Sec.~\ref{sec:problem} we seek to construct a shape servo policy of the form \(\sspolicy(\curpcloud, \goalpcloud) = \action \). We decompose our policy into two stages: (1) a feature extraction stage and (2) a deformation controller (c.f. Fig.~\ref{fig:DeformerNet} top).

The feature extractor \(\encoder(\pcloud)=\feat\) takes a point cloud as input and outputs a shape feature vector we define as \(\feat\). We use two parallel feature extraction channels taking as input \(\curpcloud\) and \(\goalpcloud\)  and generating feature vectors \(\featcur\) and \(\featgoal\) respectively. We then take the difference of these two to define the feature displacements: \(\deltafeat = \featcur - \featgoal\).

Our deformation control function, \(\deformfunc\), takes this feature displacement as input and outputs the desired instantaneous change in end-effector position, hence:
\(\action = \deformfunc(\deltafeat)\).

The composite shape servo policy thus takes the form \(\sspolicy(\curpcloud, \goalpcloud) = \deformfunc(\encoder(\curpcloud) - \encoder(\goalpcloud)) = \action \). We then use a resolved rate controller to compute the desired joint velocities following the desired end-effector displacement output by our shape servo policy \(\sspolicy\).

Training this model takes a straightforward supervised approach. We simply record the robot manipulating an object of interest, set the terminal object point cloud as \(\goalpcloud\), select any previous point cloud from the trajectory as \(\curpcloud\) and the associated end-effector displacement between the two configurations as \(\action\).
We give further details of this training procedure in Sec.~\ref{sec:experiments}.

\subsection{\DeformerNet{}  Architecture Details}
As described previously, \DeformerNet{} consists of two stages: feature extraction and deformation control. Our feature extractor uses three successive PointConv~\cite{PointConv2019} convolutional layers that successively output clouds of dimension (64, 512), (128, 256) and ultimately a 256-dimension vector that acts as the shape feature. We downsample the input current, \(\curpcloud\), and goal point clouds, \(\goalpcloud\), to 1024 points using the furthest point sampling method from~\cite{Qi2017PointNet} before inputting them into the network. We provide full details of the architecture in the bottom of Fig.~\ref{fig:DeformerNet}.

The deformation control stage takes this 256-dimension \textit{differential feature vector} and passes it through a series of fully-connected layers (128, 64, and 32 neural units, respectively). The fully-connected output layer produces the desired 3D displacement. We use ReLU activation function and group normalization~\cite{wu2018group} for all convolutional and fully-connected layers except for the linear output layer.

We use the standard mean squared error loss function for training \DeformerNet{}. We adopt the Adam optimizer and a decaying learning rate which starts at $10^{-3}$ and decreases by 1/10 every 50 epochs.

\subsection{Manipulation point prediction}\label{sec:mani_point}
As discussed above and shown in Fig.~\ref{fig:mani_point} the location at which the robot grasps the object greatly influences whether the robot can reach a target shape. As such we present here an approach to selecting an appropriate manipulation point prior to performing the shape control task.
Recall we wish to find a manipulation point on the surface of the object, \(\manippoint \in \object\). However, we must infer this location given the initial \(\initpcloud\) and target point clouds, \(\goalpcloud\) prior to acting.
We propose the use of a keypoint-based heuristic to select the manipulation point. Our preliminary work~\cite{thach2021deformernet} showed this heuristic slightly outperformed a regression-based approach.

Our heuristic follows a simple idea, points that move more should generally be closer to the manipulation point. Assume we have a set of \(K\) keypoint matches \(\matches=\{(u_j,  v_j) | u_j \in \initpcloud, v_j \in \goalpcloud\}_{j=1:K}\) between the initial and goal point cloud.
We define the associated keypoint displacements as $\delta_k = \{\|u_j-v_j\|\}_{j=1:K}$. We then estimate the manipulation point as the location defined by the displacement-weighted average of the $M$ keypoints with largest displacement.

We use an unsupervised keypoint detection algorithm based on the Transporter Network of~\cite{Tejas2019unsupervised}. The original Transporter network defines an unsupervised reconstruction loss between source and target image pairs from a video sequence.
To adapt transporters to our 3D manipulation point prediction problem, we leverage pairs of source-target point clouds collected in simulation to train the model. We convert the point cloud data to an organized, array-like point cloud format to make them compatible with the original Transporter network architecture.

\section{Experiments and results}\label{sec:experiments}
\begin{figure}[ht!]
    \centering
    \includegraphics[width=0.445\linewidth]{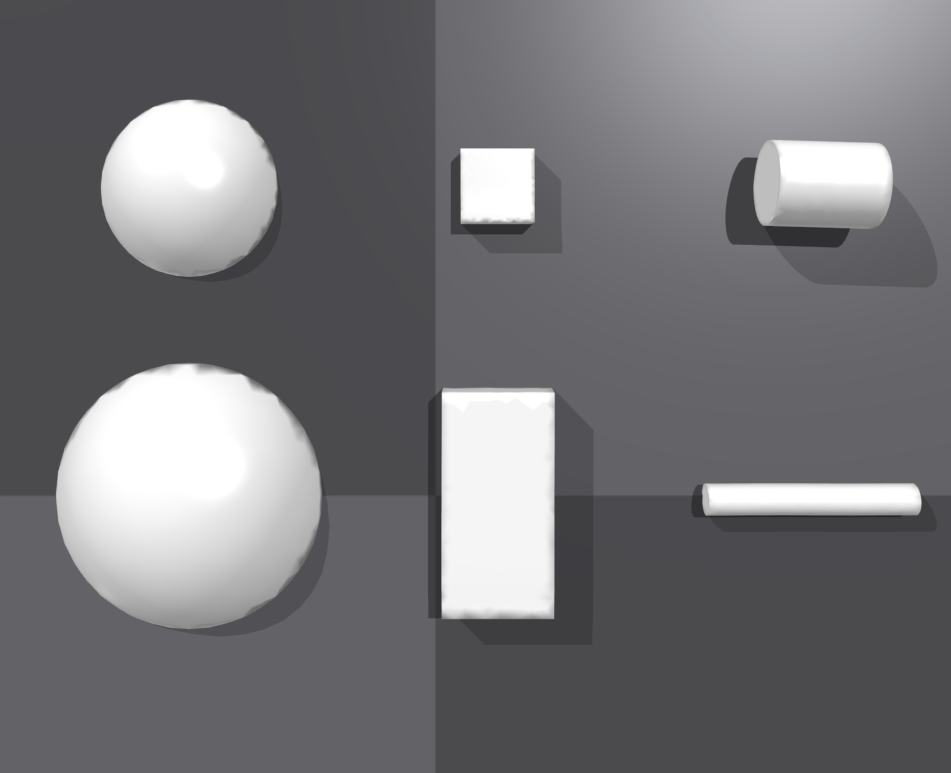}
    \includegraphics[width=0.49\linewidth]{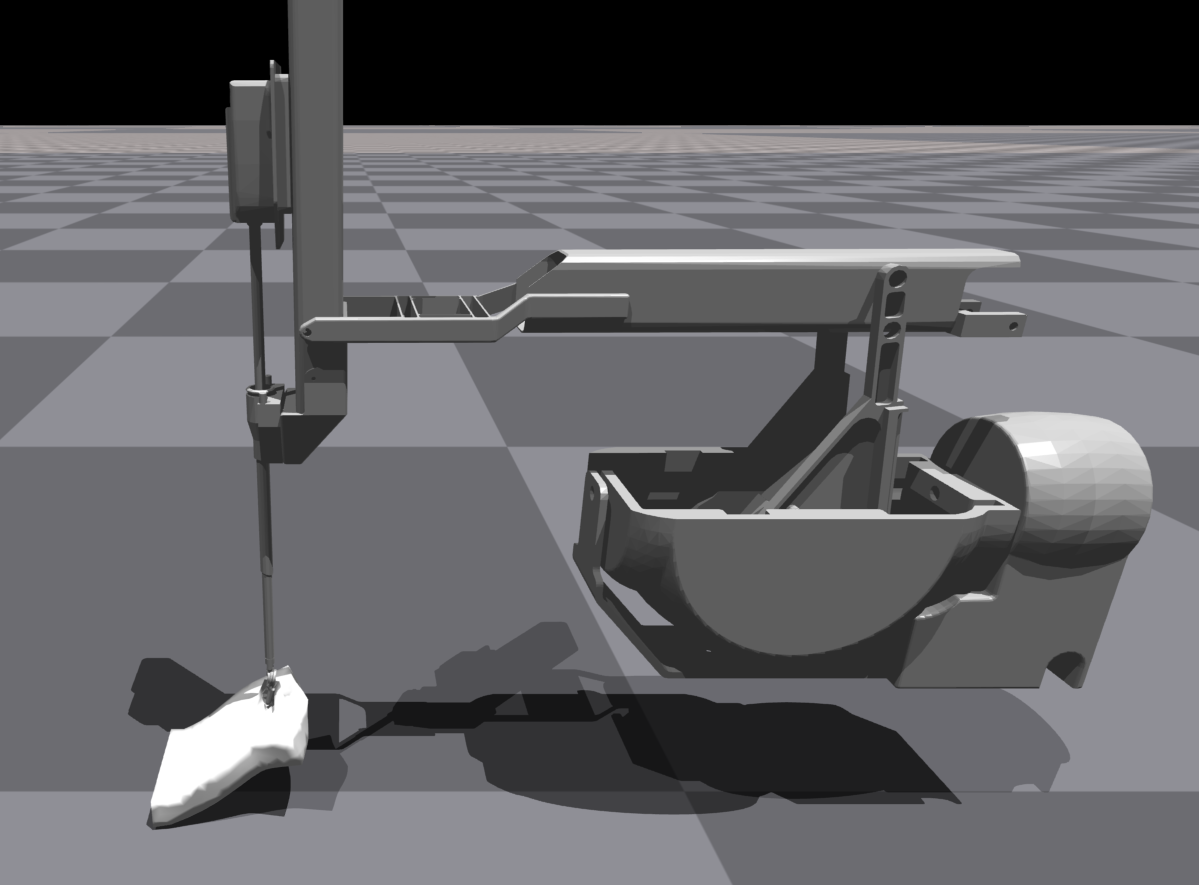}    
    \caption{(Left) We train on random interpolations of these shapes. (Right) Experimental setup showing a patient-side manipulator of the dVRK in Issac gym.}
    \label{fig:display_objects}
    \vspace{-15pt}
\end{figure}

We evaluate our method in both simulation, via the Isaac Gym environment \cite{Liang2018GPU}, and on a real robot.
For both simulation and real robot experiments, training data for the learned models are generated in Isaac Gym.
In Isaac Gym, we use a simulation of a patient-side manipulator of the daVinci research kit (dVRK)~\cite{Kazanzides2014_ICRA_DVRK} robot to manipulate objects (see Fig.~\ref{fig:display_objects} (right)).
For the real robot experiments, we use a Baxter research robot with a laparoscopic tool attached to its end effector and an Azure Kinect camera generating point clouds of the deformable object (see Fig.~\ref{fig:intro}).
In both cases, we affix one end of the deformable object to the environment and task the robot with manipulating it via one grasp point.

\subsection{Goal-Oriented Shape Servoing}
We evaluate our method's ability to deform the object to the goal point cloud.
In our previous workshop paper~\cite{thach2021deformernet}, we reported the performance of our method when the model was trained and tested on one object geometry and Young modulus and demonstrated that our method outperforms a current state-of-the-art method for learning-based 3D shape servoing by Hu \emph{et al.}~\cite{Hu20193-D}.

\subsubsection{Training Data Generation} We expand on this evaluation in this work by first evaluating our method's ability to control the shape of a variety of 3D deformable object shape primitives, including hemispheres, rectangular boxes, and cylinders (see Fig.~\ref{fig:display_objects}).
For each primitive, we investigate three different stiffness values (represented by Young's modulus): 1 kPa, 5 kPa, and 10kPa, which represent stiffness properties similar to those seen across different biological tissues~\cite{hinz2012mechanical,handorf2015tissue}.
The three shape primitives, each with three stiffness values result in a total of nine object types for evaluation.

For each of the nine object types, we create a training dataset of objects with geometries sampled uniformly at random from interpolations between the sizes of the shapes in Fig.~\ref{fig:display_objects}.
In addition, each object for training is assigned a Young modulus sampled from a Gaussian distribution with means and standard deviations of (1kPa, 0.2kPa), (5kPa, 1kPa), and (10kPa, 1kPa) for the 1 kPa, 5 kPa, and 10 kPa test scenarios, respectively. We train a separate model for each of the nine objects, using the same \DeformerNet{} architecture.

We generate each training dataset by randomly sampling 300 pairs of initial object configurations and manipulation points.
For each pair, the robot deforms the object to 10 random shapes for a total of 3000 random trajectories.
We record partial-view point clouds of the object and the robot’s end-effector positions at multiple checkpoints during the execution of this trajectory using the depth camera available inside the Issac gym environment.
We form supervised data input-output pairs for training \DeformerNet{}. The input, \((\tpcloud, \goalpcloud)\) consists of a point cloud along the trajectory at any arbitrary time \(t\), as well as the point cloud at the end of this trajectory. We compute the output, $\deltapose_t$, as the displacement between the end-effector position at time \(t\) and the end of trajectory. We sample 10,000 such pairs of data points for training our model. 

\subsubsection{Generalization Performance} We are interested in evaluating the performance of our method on test scenarios that are both inside and outside the training distributions in simulation.
To generate test scenarios outside the training distribution, we sample objects with random dimensions smaller than the minimum and larger than the maximum of each of our primitive-shaped objects. We additionally sample Young moduli with values 2-4 standard deviations from the mean of the training distribution moduli.
For each test scenario we select 10 random objects from inside the training distribution and 10 from outside the training distribution. We then sample 10 random goal shapes for each of the 20 test objects. We select the manipulation point for testing using our keypoint-based heuristic with \(K = 200\) keypoints.

\begin{figure}[ht!]
    \centering
    \vspace{-6pt}
    \includegraphics[width=0.95\linewidth,clip,trim=1.3cm 0.2cm 2.6cm 0.8cm]{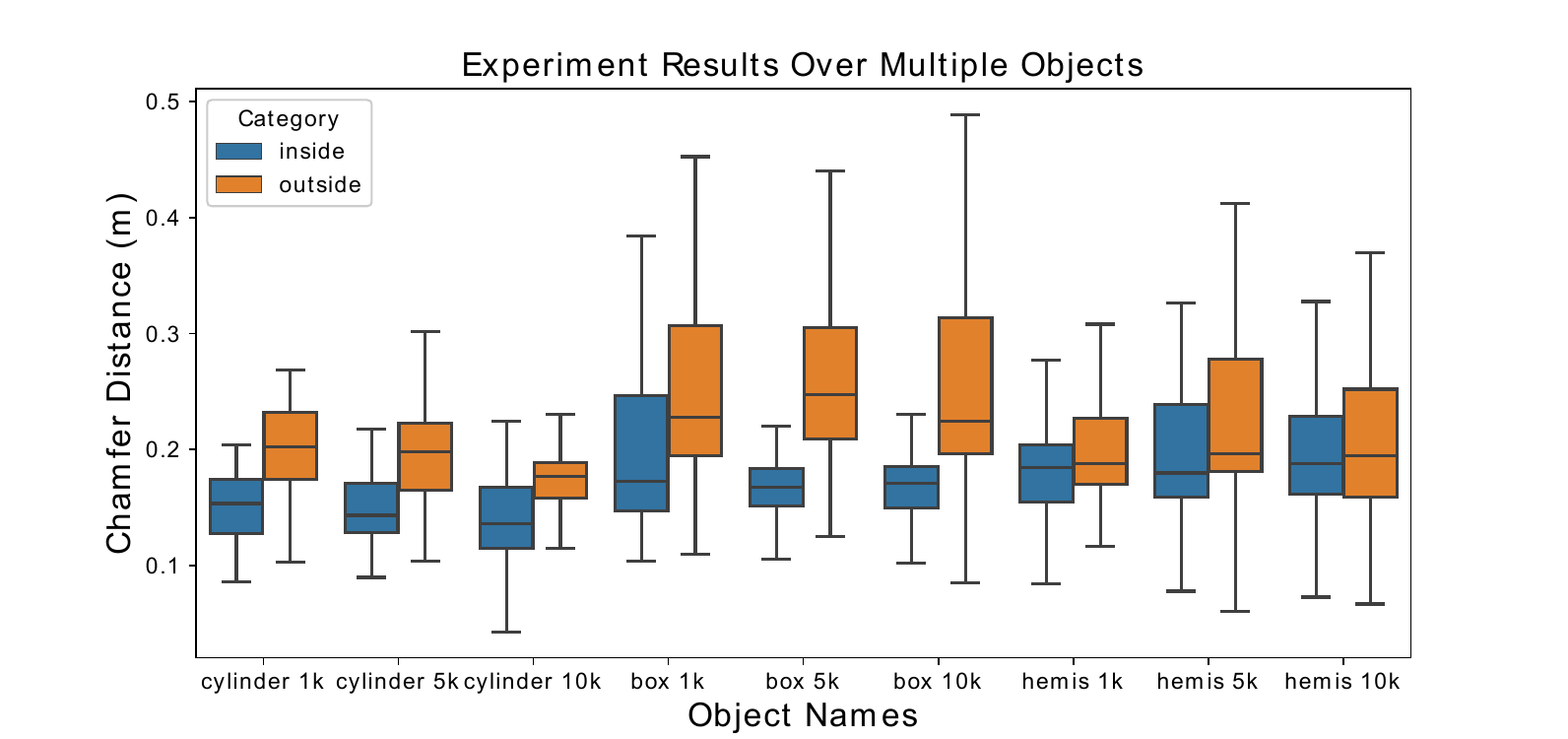} %
    \captionsetup{skip=3pt}
    \caption{Distribution of Chamfer distance after shape servoing. ``Inside'' reports for object inside the training distribution and ``outside'' for random objects outside the training distribution.}
    \label{fig:chamfer1}
    \vspace{-20pt}
\end{figure}

\begin{figure*}[hbt!]
    \centering
    \includegraphics[width=1.0\textwidth,clip,trim=0cm 3cm 0cm 0.5cm]{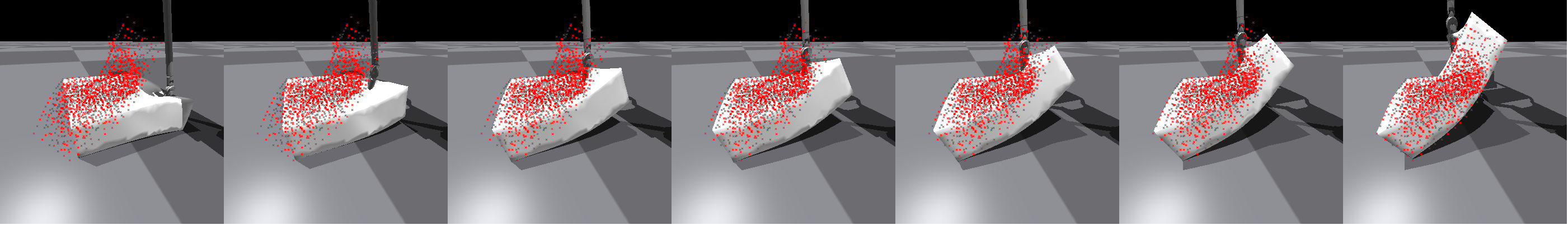}
    \includegraphics[width=1.0\textwidth]{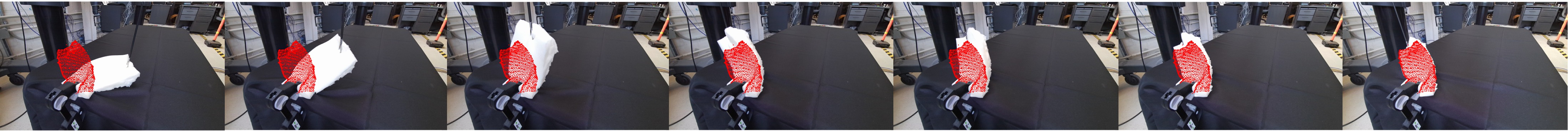}
    \includegraphics[width=1.0\textwidth]{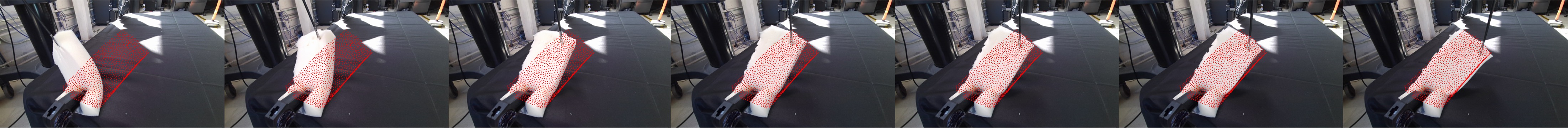}    
    \caption{Sample manipulation sequences of DeformerNet in different setups. The sparse red clouds visualize the target shapes of the object. First row: with simulated dVRK in Isaac Gym (0.18 m final Chamfer dist.). Second row: with physical robot and real goal point clouds (0.30m final Chamfer dist.). Third row: with physical robot and simulated goal point clouds (0.39m final Chamfer dist.).}
    \label{fig:sequence1}
    \vspace{-12pt}
\end{figure*}

We use Chamfer distance as our primary evaluation metric to describe how close the final manipulated object's point cloud is to the goal point cloud.
Chamfer distance computes the average distance of each point in one point cloud to the closest point in the other point cloud, \(d_c = \frac{1}{|\mathcal{P}_1|}\sum_{x\in \mathcal{P}_1}\min_{y\in \mathcal{P}_2} ||x - y||^2\).
Fig.~\ref{fig:chamfer1} visualizes the result of each object type with a boxplot recorded over the 20 test objects with 10 goal shapes each. The box represents the quartiles, the center line the median, and the whiskers represent min and max final Chamfer distance.
For visualization purpose, we also provide a sample snapshot of the robot performing shape servoing to a goal shape in Fig.~\ref{fig:sequence1}.

The experiment results show that our method is capable of generalizing what it learns from training to adapt to geometries, material properties, and goal shapes it has never seen before, both inside and outside the training distribution, although predictably with some fall off in performance outside the training distribution. We observe that the primary common cause of failure comes from the heuristic manipulation point predictor selecting a grasp location unable to achieve the goal shape.   

\subsubsection{Baseline Comparisons}
We also compare the performance of our method against Rapidly-exploring Random Tree (RRT) \cite{lavalle2001randomized} and model-free Reinforcement Learning (RL) for the 3D shape servo problem.
Here we restrict the task to be trained and tested on a single box object as described in~\cite{thach2021deformernet} and use only one manipulation point throughout training and testing.

For the RRT implementation, we define the configuration space as the joint angles of the dVRK manipulator.
We define a goal region as any object point cloud that has Chamfer distance less than some tolerance from the goal point cloud.
We use the finite element analysis model~\cite{macklin2019} in the Isaac Gym~\cite{Liang2018GPU} simulator to derive the forward model for RRT.

We use proximal policy optimization (PPO)~\cite{schulman2017proximal} (as in~\cite{pore2021safe}) with hindsight experience replay (HER)~\cite{andrychowicz2017hindsight} for model-free RL.
We use our \DeformerNet{} architecture for the actor and critic network except for the critic output being set to single scalar to encode the value function.
Each episode we condition the policy on a newly sampled goal shape.
We train the RL agent with 100,000 samples---10 times the amount of data provided to \DeformerNet{}.

\begin{figure}[th]
    \centering
    \vspace{-6pt}
    \includegraphics[width=0.8\linewidth,clip,trim=0.2cm 0.0cm 0cm .8cm]{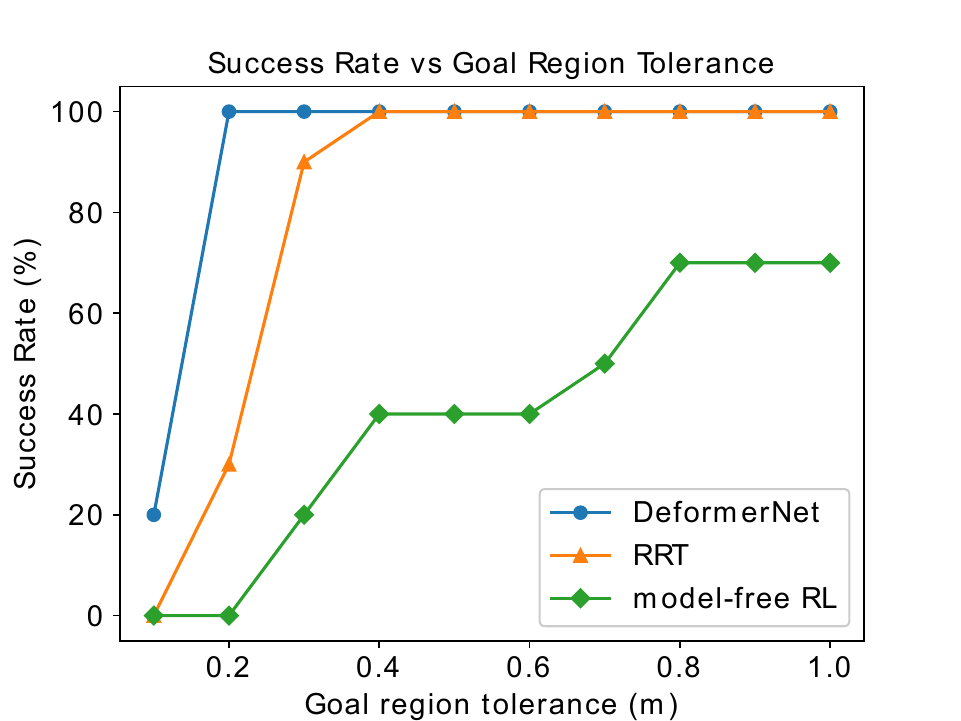}
    \caption{Success rate comparison of \DeformerNet{} to RRT and RL baselines for varying levels of goal tolerance.}
    \label{fig:success_rate}
    \vspace{-18pt}
\end{figure}

We evaluate DeformerNet, RRT, and model-free RL with 10 random goal shapes.
Fig.~\ref{fig:success_rate} shows the success rate of the three methods at different levels of goal tolerance.
We clearly see that even with 10 times the training data compared to our method, the model-free RL agent achieves a significantly lower success rate compared to the other two methods.
We also note that while RRT succeeds comparably to our method at looser goal tolerances, at tighter goal tolerances RRT fails more often.
Further, unlike our method, RRT does not incorporate feedback during execution. As such RRT will not be able to recover if the object shape deviates from the plan. While one might think to perform replanning, we note that RRT requires several orders of magnitude more computation time required than our shape servoing approach.
For instance, at a tolerance of 0.4 (where both our method and RRT achieve 100\% success), over the 10 test goal shapes, the lowest computation time required by RRT was 3.3 minutes, the highest was 121.6 minutes, mean was 38.7 minutes, and standard deviation was 40.1 minutes. Our DeformerNet, however, only requires a pass through the neural network which takes minimal time.
As a result, for this task, we note a significant success rate improvement for our method over model-free RL, a success rate improvement at strict goal tolerance values over RRT, and a significant computation time improvement over RRT in all cases.
\begin{figure*}[t!]
  \centering
    \includegraphics[width=1.0\textwidth]{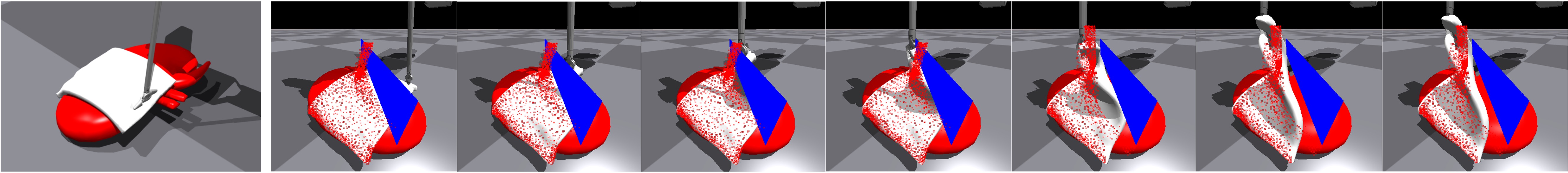} \\
    \includegraphics[width=1.0\textwidth]{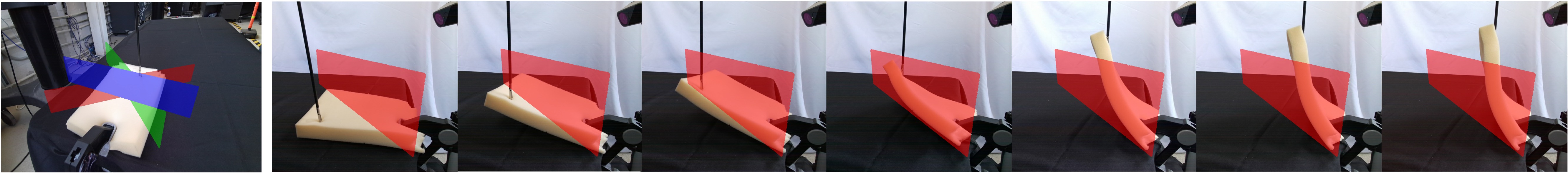}    
    \caption{\textit{Top row}: simulated retraction experiment setup (left) and a sample successful retraction sequence with target plane visualized in blue. \textit{Bottom row}: visualization of target planes for physical robot retraction (left) and a successful sequence with target plane in red.}
    \label{fig:sequence_plane}
    \vspace{-12pt}
\end{figure*}

\subsubsection{Physical Robot Goal-Oriented Shape Servoing}
We next evaluate our method's ability to perform shape servoing on the real robot, while having been trained entirely in simulation.
The experimental setup (shown in Fig.~\ref{fig:intro}) leverages a foam box affixed on one side to a table.
We segment the object's point cloud out from the rest of the scene by fitting a plane to the table with RANSAC~\cite{fischler_bolles_1981} and selecting the points above this planes.
We filter out the black table clamp and the laparoscopic tool using pixel intensity. 

\begin{figure}[th]
    \centering
    \vspace{-8pt}
    \includegraphics[width=0.8\linewidth,clip,trim=0.2cm 0cm 0cm 0cm]{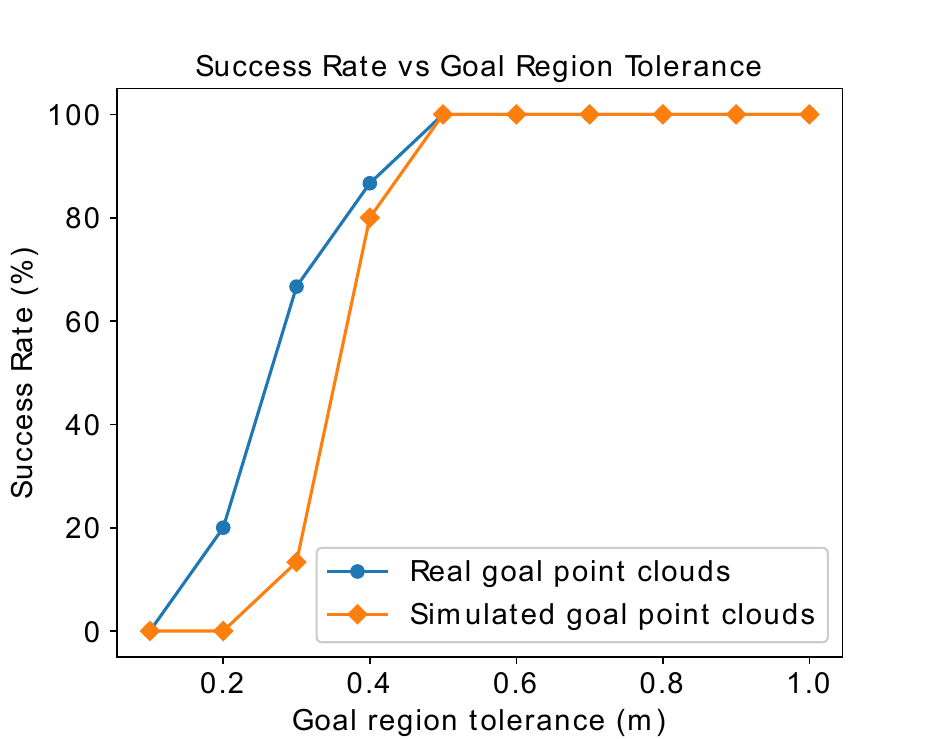}
    \caption{Success rate of \DeformerNet{} with physical robot when given real goal point clouds and simulated goal point clouds.}
    \label{fig:success_rate_real}
    \vspace{-6pt}
\end{figure}

\begin{figure}[ht]
  \centering
  \includegraphics[width=0.49\linewidth]{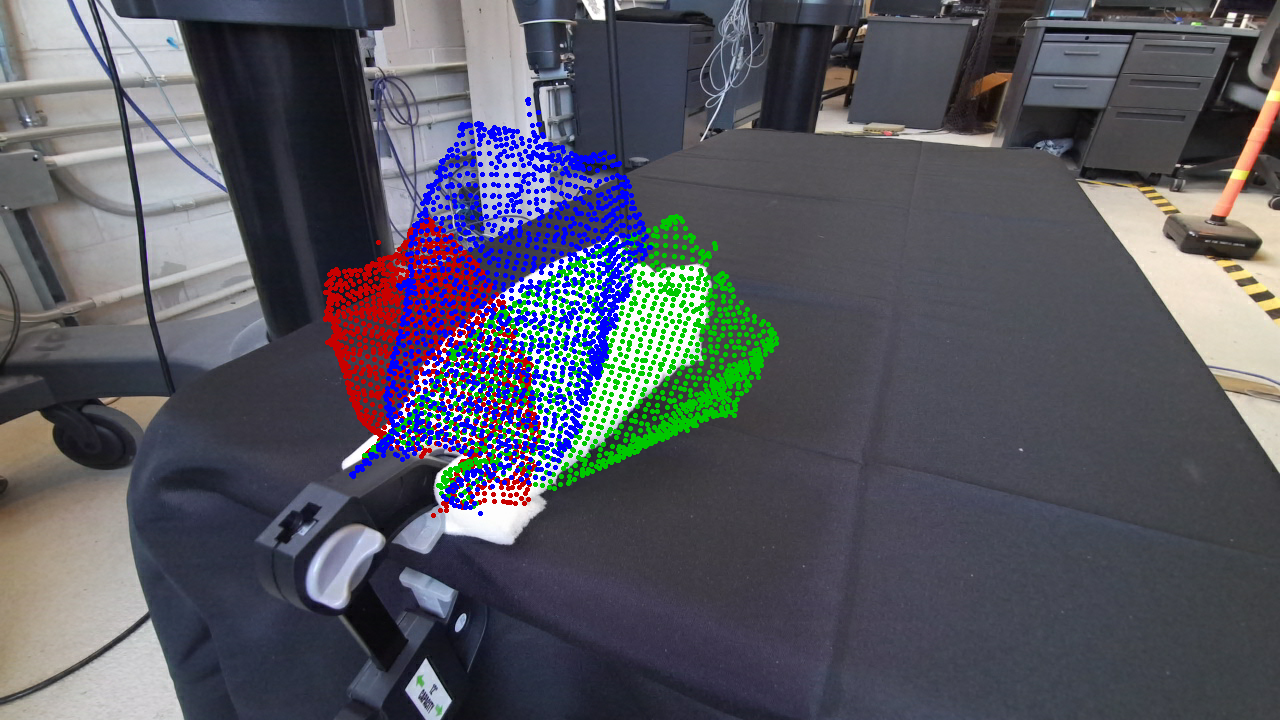}
  \includegraphics[width=0.49\linewidth]{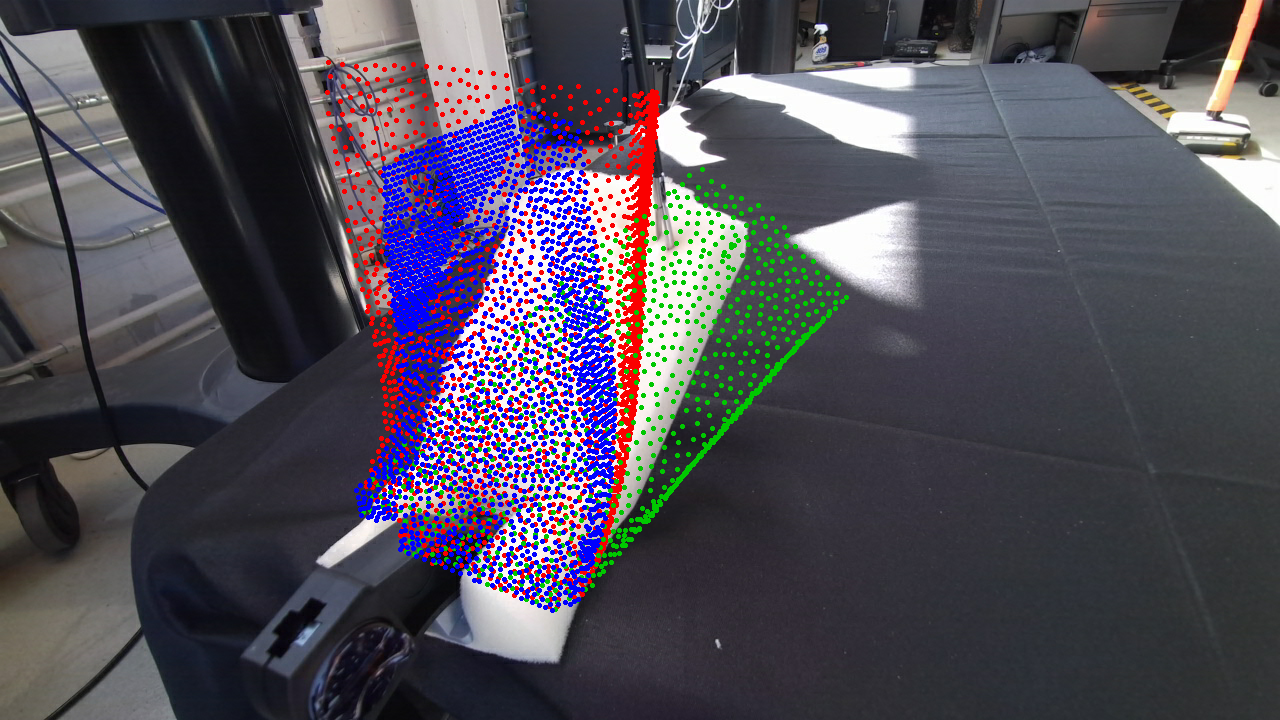}
    \caption{Left: goal point clouds from real sensor recordings. Right: goal point clouds generated in simulation.}
    \label{fig:all_goals}
    \vspace{-20pt}
\end{figure}

We generate three distinct goal shapes (Fig.~\ref{fig:all_goals} (left)) by manually moving the object to random shapes with the laparoscopic tool and recording the resulting point cloud. Figure~\ref{fig:success_rate_real} describes the success rate of the 15 trials over different goal tolerance levels.
Figure~\ref{fig:sequence1} visualizes a typical manipulation sequence.

To showcase our method's robustness, we additionally evaluate on 3 goal point clouds obtained entirely from the simulator (Fig.~\ref{fig:all_goals} (right)). Figure~\ref{fig:success_rate_real} visualizes the success rate of the 15 trials over different goal tolerance levels.
A sample visualization is provided in Fig.~\ref{fig:sequence1}. Overall we note a slight drop in quantitative performance in the real world compared to simulation, while qualitatively still succeeding.

\subsection{Surgical Retraction}
We next evaluate our method's ability to perform a mock surgical retraction task, in which a thin layer of tissue is positioned on top of a kidney.
We task the robot with grasping the tissue layer and lifting it up to expose the underlying area.
Figure~\ref{fig:sequence_plane} (top, left) shows the simulation environment composed of a kidney model with a deformable tissue layer placed over it and fixed to the kidney on one side. We train \DeformerNet{} on a box object similar in dimensions to the tissue layer, but without the kidney present. 

Instead of requiring the operator (e.g. surgeon) to provide an explicit shape for the robot to servo the tissue to, we instead just require them to define a plane which the tissue should be folded to one side of. An example plane can be seen in Fig.~\ref{fig:sequence_plane}.
We use a simple algorithm to infer a goal point cloud for the object based on this target plane. We use RANSAC~\cite{fischler_bolles_1981} to find a dominant plane in the object cloud and then find the minimum rotation to align this plane with the target plane. We then apply this estimated transformation to any points not lying on the correct side of the plane and set this as the target cloud along with the points currently satisfying the goal. If after reaching the goal point cloud any part of the object still resides on the wrong side of the plane, we shift the target plane further into the goal region along the plane's normal vector and repeat the entire process.

To evaluate we sample 100 random planes with differing orientations in simulation and task the method with moving the tissue layer beyond the plane. Our approach reveals the kidney underneath with a success rate of 95\%.

We also evaluate retraction on the physical robot.
We affix a thin layer of foam to the table and task the robot with moving the object via the laparoscopic tool beyond a target plane.
We evaluate on 3 different planes (see Fig.~\ref{fig:sequence_plane}), and for each plane conduct 5 trials.
We observe a 100\% success rate across the 15 trials.
We provide visualizations of representative retraction experiments in Fig.~\ref{fig:sequence_plane}.

\section{Conclusions}\label{sec:conclusions}
In this paper we presented a novel-approach to closed-loop 3D deformable object shape control. Crucially we demonstrate through rigorous simulated and physical-robot experiments that shape servoing with \DeformerNet{} can manipulate objects with novel material properties or shape while only requiring a partial-view 3D point cloud as input. We further demonstrate how our shape servoing approach can be adapted to the task of surgical retraction, where a much simpler goal representation in the form of a separating plane needs only be provided. Our future work aims to extend our manipulation approach to more surgical tasks, to manipulation of plastic materials~\cite{huang2020plasticinelab}, and manipulating deformable 3D objects common to homes and warehouses. Finally, we wish to move beyond our greedy, visual servoing approach to provide more explicit planning for longer-horizon tasks.

\bibliographystyle{IEEEtran}
\bibliography{bibliography}

\end{document}